%% file: Main_Final_Arxiv.tex
\documentclass[journal,single]{IEEEtran}
\ifCLASSINFOpdf

\else

\fi

\usepackage{amsmath}
\usepackage{color,soul}
\usepackage{graphicx}
\usepackage{makecell,tabularx}

 \def \et {\it et al.}
\begin{document}

\title{Deep Neural Network Perception Models and Robust  Autonomous Driving Systems}

\author{Mohammad Javad Shafiee,~\IEEEmembership{Member,~IEEE,}
        Ahmadreza Jeddi,
        Amir Nazemi,
        Paul~Fieguth~\IEEEmembership{Senior~Member,~IEEE,}
        and~Alexander~Wong~\IEEEmembership{Senior~Member,~IEEE} 
\thanks{M.~J. Shafiee, A. Nazemi, P. Fieguth, and A. Wong are with the Waterloo Artificial Intelligence Institute and the Department
of Systems Design Engineering, University of Waterloo. A. Jeddi is with the David Cheriton school of Computer science, University of Waterloo,  Waterloo,
ON, Canada.}
}


\maketitle

\IEEEpeerreviewmaketitle

\section{Introduction}

The National Highway Traffic Safety Administration (NHTSA) reported that over 90\% of in-road accidents in 2015 occurred purely because of drivers' errors and misjudgements, with factors such as fatigue and other sorts of distractions being the main cause of these accidents~\cite{singh2015critical}. One promising solution for reducing (or even resolving) such human errors is via autonomous or computer-assisted driving systems.  As such, autonomous vehicles (AVs) are currently being designed with the aim of reducing fatalities in accidents by being insusceptible to typical driver errors. Moreover, in addition to improved safety, autonomous systems offer many other potential benefits to society: i) improved fuel efficiency beyond that of human driving, making driving more cost-beneficial and environmentally friendly,  ii) reducing commute times due to improved driving behaviours and coordination amongst autonomous vehicles, and iii) better driving experience for individuals with disabilities, to name a few.  

Given the extensive global interest towards the deployment of AV technologies, recent studies have introduced new guidelines and regulations for speeding up AV development and pushing AVs into the market in a more effective manner. At the same time, there have been significant efforts to inform the public on the  capabilities  and limitations of AV systems. 

The most widely used approach to categorize  AV systems is to classify them based on their level of automation, as standardized by Society of Automotive Engineers (SAE)~\cite{sae2016taxonomy}, ranging from level zero (no automation) to level five (completely autonomous). Although a level five AV is the ideal, the majority of current AV systems are at levels one and two.  The limited autonomous driving capabilities of current AV systems are due to a range of challenges such as the high cost of sensors, a lack of acceptance by the public, a lack of appropriate safety evaluations, and the high error rates of existing technologies. In this study, we focus on those challenges associated with level three or higher~\cite{sae2016taxonomy}.

\begin{figure*}
    \centering
    \includegraphics[width=1\textwidth]{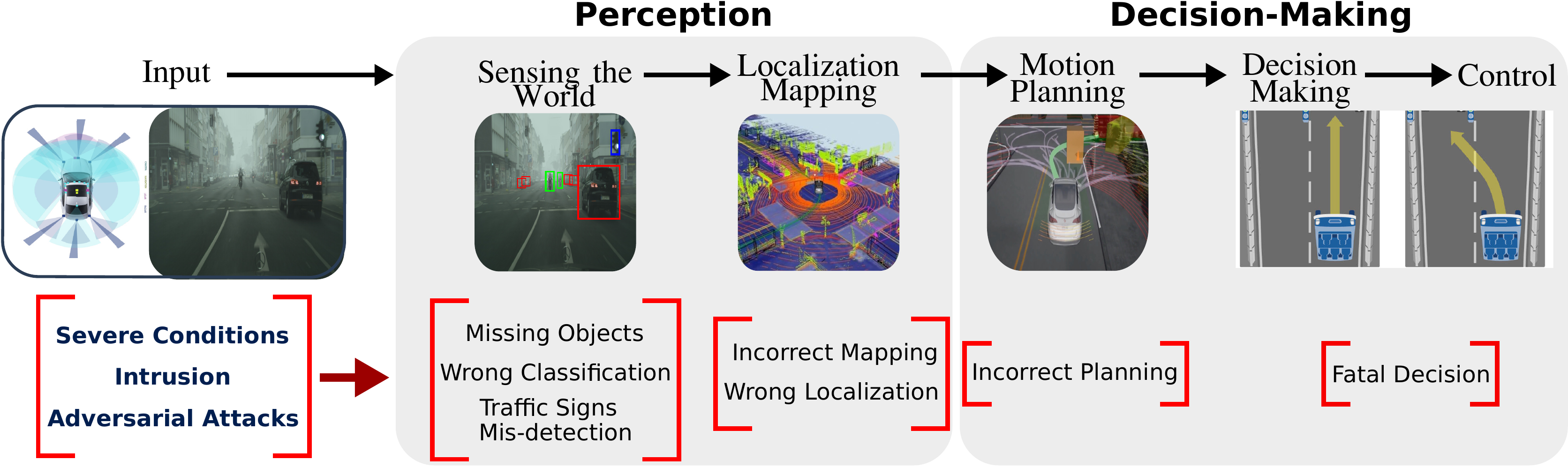}
    \caption{The general architecture of an autonomous driving system, which is comprised of two  main components of Perception and Decision-making. The perception system typically consists of a set of machine learning algorithms, and provides a semantic understanding of the world around the vehicle. The perception system can be negatively affected by different types of intrusion algorithms, commonly referred to as \textit{adversarial attacks}.  The perception system (and especially the `Sensing the World' module) are the first step between the outside world and the autonomous driving system; therefore, any incorrect conclusions from the perception system, due to adversarial attacks, will propagate to later components, leading to potentially fatal decisions being made. The Decision-Making component of an AVs is comprised of Motion Planning, Decision Making and Control. These modules are responsible to identify the best path for the vehicle and to actuate the car towards that.}
    \label{fig:AV_flowDiagram}
\end{figure*}

Although different levels of automation can lead to some variations in the developed systems; the general architecture of an autonomous driving system consists of five main components, as shown in Figure~\ref{fig:AV_flowDiagram}, grouped into  two main areas of I) perception and II) decision/control. Perception includes all of the hardware and software attempting to find the current state of the AV system with respect to its surrounding environment, such that this information can be used as the input to decision and control. Sensors, algorithms to process the sensed data, environmental mapping, and localizing the AV with respect to the generated map are all components within perception.

Sensor devices which are typically used in AV perception include LiDAR, various types of visual cameras, GPS, RADAR, as well as  Internet of Vehicles (IoV) devices. 

Raw sensory data are processed by a variety of algorithms to generate useful information regarding the environment around the AV, of which three examples include 
\begin{enumerate}
\item {\bf Object detection}, responsible for taking sensor data and detecting important objects of interest such as traffic lights, traffic signs, road, lanes, pedestrians, and other vehicles;
\item {\bf Semantic segmentation}, responsible for segmenting the road and its participants from sidewalk and objects that are not within the road; 
\item {\bf Scene reconstruction}, responsible for generating 3D scenes based on 2D images and/or LiDAR devices.
\end{enumerate}
The perception algorithms employed are clearly a function of the automated driving level~\cite{sae2016taxonomy}, for example while cruise control (based on LiDAR or Radar sensors) is a functionality in level 1, lane-centering (visible images and a segmentation model) is one of the main tasks in level 2.  Level 3 driving systems are ones that can truly be considered autonomous, a level of autonomy which can allow drivers to sit back and relax. As such, several models are being used to perform the automated driving. Traffic signs and traffic lights are detected by object detection models, typically based on  visible camera images; other cars driving in the road can be detected using object detection models by fusing sensory data from different sources; and the valid area in which the car can drive is identified by segmentation models. While drivers can be ``hands-off" in level 3 vehicles, they need to be ready to engage and take control at any time. In contrast, drivers can be ``mind-off" in level 4 vehicles, but only in certain, geofenced traffic areas, or in Level 5 vehicles in any situation.

Different perception models\footnote{Some state-of-the-art comparison results on two different computer vision applications can be found in the supplementary material. } may be used independently for different tasks such as lane segmentation, traffic sign identification, and traffic light detection; however, this information should be fused for path planning and decision-making, for example to be able to associate different signs and lights to a particular lane in the road. Nonetheless, such fusion is still a challenging task, and as such using predefined maps can mitigate this problem to some extent in a practical scenario and increases association accuracy.

Localization and mapping systems provide a map of the environment surrounding the autonomous vehicle and determine the current state (i.e., position and orientation) of the AV with respect to this detailed map.

Given the mapping of the environment and the AV's current state, the decision system makes decisions on what actions to be taken in order to optimally reach a goal; that is, responsible for generating route proposals, motion planning for each of those routes, evaluating compliance with passenger preferences and the law~\cite{mcallister2017concrete}, calculating safety probabilities of each proposed trajectory, and making decisions on which trajectory to select.  The control system acts upon these decisions and controls the vehicle, generating appropriate commands and ensuring that the proposed trajectory is followed.
There remain several challenges regarding the interaction of these two parts.  In particular, the decision-control system should reevaluate the possible risks in different situations constantly and predict the intentions of human drivers around the vehicle. As such, effectively estimating uncertainty is very important, however  understanding human driver intention is still not a common practice in the field, and is usually relaxed  in the problem formulation.

A more granular description of the five components of an autonomous driving system is as follows:
\begin{enumerate}
    \item {\bf Sensing the world} consists of various sensors and algorithms processing the available data and provides a semantic scene understanding~\cite{cheng2011autonomous}.
    
    \item {\bf Localization and mapping} computes the AV pose (location and orientation) with respect to the surrounding environment, which is frequently addressed via simultaneous localization and mapping (SLAM) techniques~\cite{dissanayake2001solution}, well-known in robotics and off-line pre-mapping.
    
    \item {\bf Motion planning} provides different trajectories as sequences of states, given the environment information, initial states, and the final goal~\cite{frazzoli2002real}. 
    
    \item {\bf Decision-making} selects the optimal trajectory, while considering other factors like safety, compliance, etc.~\cite{paden2016survey}.
    
    \item {\bf Control} actuates components and ensures that the AV follows the selected path~\cite{paden2016survey}.
\end{enumerate}
The preceding modular structure requires harmonizing all components together, training all components together, and reducing the propagation of erroneous decision-making from level to level.  As a result of these limitations, end-to-end autonomous driving systems have been proposed, in which the components are learned as a single system.  For example, Bojarski {\it et al.}~\cite{bojarski2016end} proposed PilotNet, a deep CNN based framework that takes visual data from cameras as sensory input and delivers a simple autonomous system for lane following by outputting steering angles based on a fully end-to-end approach. Caltagirone \mbox{\it et al.}~\cite{caltagirone2017lidar} utilized a fully convolutional network to generate path proposals from LiDAR sensory data, a partial end-to-end paradigm.

Whether fully end-to-end, partial, or broken into classical components, deep learning methods are now at the heart of essentially all AV technologies. Although deep neural network models provide the state-of-the-art performance in most scene understanding algorithmic tasks, the robustness of neural network models  has become a major concern in the research community.  In the context of AV systems, extreme weather conditions and possible intrusion attacks by adversaries are two key situations where these models may be particularly vulnerable.

As seen in Figure~\ref{fig:AV_flowDiagram}, incorrect scene understanding by the `Sensing the world' module, the first module in the sequence, can propagate wrong information through the pipeline of consecutive modules and end up with an incorrect decision, possibly a fatal outcome.  False information may be generated for a variety of reasons, such as severe weather conditions,  complicated urban scenarios, and intrusion by adversaries, raising major concerns on the safety and security of DNN models.  

In this paper, we study how adversarial attacks can upend the claimed or assessed robustness of deep neural networks, specifically in the application to autonomous vehicles.  We mainly discuss how these attacks can impose danger to AV perception, and how these threats can propagate through the entire pipeline.
We will analyze the existing mechanisms to address these issues, and will describe the best practices to improve AV systems.

\section{Robustness Challenges in Autonomous Driving Systems }
In this section, we describe two main factors which can influence the robustness of machine learning models specifically for autonomous driving applications. While several different criteria can affect the robustness of a machine learning model, here we focus on adverse conditions and adversarial attacks as prominent factors.
Autonomous navigation requires an understanding of the environment around the car, and machine learning models play an important role in fulfilling this task.  In particular, these models need to perform perfectly in different environments and scenarios, a degree of generalization which will be an important factor in reliable autonomous systems.

\subsection{Wide Variations and Adverse Conditions }

Generalization can be defined as the ability of a trained model to deal with samples which were not seen during training. In contrast to generalization, models can also be subject to overfitting, whereby a model essentially memorizes the training samples, and indeed performs with high accuracy on that training data, but providing poor performance on unseen  testing data. 
To measure the generalization of a model, the generalization error is defined as the difference between the expected and empirical errors. The empirical error is defined as the model error on the available sample data, whereas the expected error measures how the model can perform over variation of the data based on their true (but normally unknown) underlying distribution.
The available test set should be large enough to be able to calculate a reliable empirical error in order to quantify model generalization.

Understanding generalization is key in autonomous driving systems because of the safety-critical aspects of these systems. 
A generalized autonomous system should perform reliably in a wide variety of different conditions and variations, particularly challenging in autonomous driving applications due to the extremely stochastic environments around the vehicle, which can cause a distribution drift. For example, an autonomous vehicle trained in one country and tested in another one will face inputs that did not exist in training. 

Generalization and robustness issues have certainly been studied for different applications.  Visual domain adaptation and Generative adversarial networks (GAN) are two different techniques proposed to improve model generalization. Tzeng~{\it et al.}~\cite{tzeng2017adversarial} took advantage of GAN methods to generate new samples and they used discriminative modelling and weight sharing to improve the domain adaptation on different classification tasks.  Yang {\it et al.}~\cite{yang2018unsupervised} proposed a method to keep the generalization ability of an autonomous driving system by mapping real data into a unified domain and to make decisions on the virtual data. 

Any autonomous driving system needs to be functional in real-world contexts and outdoor environments, performing reliably in different weather conditions or different lighting situations. 

A first simple but important situation is the system performance at nighttime, particularly the functionality of vision-based models. Taking advantage of Lidar or far-infrared (FIR) sensors and data fusion is one approach, however there is a body of research \cite{dai2018dark} on directly improving the robustness of vision-based models in dark environments. 

Dai and Van Gool~\cite{dai2018dark} proposed a new adaptation mechanism to address the semantic segmentation problem at nighttime. They utilized twilight images as an intermediate step to adapt the models to darker environments before fine-tuning the model for nighttime scenes, a so-called Gradual Model Adaptation process.

The main issue in generalizing models to adverse conditions is the availability of training data to use in model learning. For example, thick fog is observable only during 0.01\% of typical driving in North America; therefore, having enough samples for annotation and preparing training data is very challenging. 
As a result, generating synthetic data may be highly desirable.  Sakaridis {\it et al.}~\cite{sakaridis2018semantic} introduced a new approach to synthesize foggy driving scene data from clear-weather outdoor scenes; their results showed that utilizing synthetic data helps the model to generalize better on foggy scenarios and leads to more robust predictions.  

While providing enough data with sufficiently varied/adverse conditions is necessary, devising proper frameworks and method to handle these conditions are crucial as well. Utilizing LiDAR or FIR data might resolve nighttime scenarios, however acquiring reliable LiDAR information might be challenging in severe conditions such as rain, snow or foggy situations. As a result, fusing information from different sources and sensors is a common approach. Several methods have focused on when and in which step such fusion should be performed. Methods can be divided into two main streams of early-fusion~\cite{bijelic2019seeing}, where the features extracted from each sensor are intertwined and knowledge fusion is performed in an early stage of the network, versus late-fusion~\cite{ku2018joint}, where each sensor datum is processed independently and the results are combined at the end.

\subsection{Intruding Autonomous Driving Systems }

Besides the natural challenges just discussed, unnatural factors may challenge autonomous driving systems as well.  Essentially every software system is prone to some sort of intrusion, whether via physical access to the system or intrusions without any explicit interaction between intruder and physical system.  For autonomous vehicle systems, it is the machine learning models which are our main concern in this paper; since learned models understand the world based on sensory data,  it is possible to intrude the system by providing deceiving sensory input which fools the machine learning models.

\subsubsection{Adversarial Attacks}
Generally, attacks can be applied on any part of an intelligent system, such as on the training data (training set poisoning), model output (model theft), and manipulated inputs (adversarial examples).  Since careful model learning should avoid or limit the first two effects, it is the most likely attack, that of manipulating sensor inputs as adversarial attacks, which will be our focus.

Adversarial examples are those input samples that can fool a trained model with a high confidence. In particular, those input samples which reliably fool the trained model, at the same time only barely (or not at all) perceptible to the human eye.  In an adversarial attack, the main goal is to find an input sample $x^\prime$ close to the true sample $x$ which changes the prediction from conclusion $y = f(x)$ to conclusion $y^\prime = f(x^\prime)$, such that $y \neq y^\prime$. ideally the difference between $x$ and $x^\prime$, the perturbation size, is very small, making the input perturbation close to imperceptible.  Typically the perturbation is measured using an $l_p$ norm:
\begin{equation}
    \eta = \|x^\prime - x \|_p.
\end{equation}
There are two types of adversarial attacks, \text{\em white-box} attacks and \text{\em black-box} attacks \cite{chakraborty2018adversarial}. In a \text{\em white-box} attack, the adversary has a full access to the trained model, and knowledge regarding the model structure and parameters to allow for a fairly informed attack.  In a \text{\em black-box} attack the adversarial method does not have access to the model, so the adversary has to query the target model in order to estimate those needed aspects of the model's interior structure. 
 
Fast Gradient Sign Method (FGSM)~\cite{goodfellow2014explaining} and DeepFool~\cite{moosavi2017universal} are two simple, but well-known, examples of white-box attacks. These two attacks fall into the class of first-order adversaries, which use the gradients of the network loss function with respect to the input data to perturb input samples into adversarial examples. 

However black-box attacks are more applicable in real world scenarios, and certainly more applicable in autonomous driving systems, as these systems are not accessible, in general, to outsiders.  Black-box attacks may be undertaken by ensemble-based approaches, which use multiple trained models and generate common adversarial examples, which are then validated on the target model.  In other words,  ensemble-based attacks generate adversarial samples using a \text{\em white-box} attack, which are then utilized to attack the target model in black box form, what is known as a transferable attack.  A second black-box approach is that of Zeroth-Order Optimization~\cite{chen2017zoo}, which tries to estimate the gradient and Hessian of the network function using the inputs and outputs of the model.

Another perspective is to divide adversarial methods into \text{\em targeted} and \text{\em non-targeted} attacks. A \text{\em targeted} approach tries to change the input in a way to have the model predicts a particular specified (\text{\em targeted}) class label.  For example, inducing the network to recognize a stop sign as a speed limit sign could be the outcome of this method. A \text{\em non-targeted} method manipulates a trained model to misclassify the input data without constraining the objective toward any specific class label.

Furthermore, for applications like object detection there are positive and negative classes, and adversarial samples are generated to either fool the model to not to detect the object, or to classify the object to a wrong class label. For example, a vision based autonomous driving system may mis-classify a tree as a traffic sign (false positive) or may not classify a traffic sign at all (false negative).  Table~\ref{tab:adversarial_attack_summary} summarizes the most well-known adversarial attack algorithms.

\begin{table*}[]
    \centering
    \caption{Summary characteristics of common adversarial attacks.  Attacks are characterized based on whether they are white-box or black-box (i.e., having accessing to model details or not), the number of network queries to be able to generate the perturbed input (attack frequency), and how they measure the amount of perturbation added to the input data. These characteristics are described based on the most common approaches which these methods are used, however it is possible to extend these methods to change the characteristics as well. More details and the references to these methods can be found in the supplementary material~\cite{}. }
    \begin{tabular}{|l|c|c|c|c|}
    \hline
        \bf Method &\bf White-Box/Black-Box & \bf Targeted/Non-Targeted & \bf Attack Frequency  &\bf  Measurement  \\ \hline
        FGSM 
        &  White-Box  & Non-Targeted  &     One-time  & element-wise \\ \hline
        DeepFool 
        &  White-Box & Non-Targeted   &    Iterative &   $l_p$  \\ \hline
        C\&W 
        &  White-Box    & Targeted    &Iterative &  $l_1,l_2,l_\infty$   \\ \hline
        PGD 
        & White-Box       &   Both  &Iterative &   $l_\infty$  \\\hline
        ZOO 
        & Black-Box & Both   &Iterative &    $l_2$ \\ \hline
        One-Pixel 
        & Black-Box      & Both    &Iterative &   $l_0$  \\ \hline
    \end{tabular}
    \label{tab:adversarial_attack_summary}
\end{table*}

\subsubsection{Defence Mechanisms}

In addition to the development of algorithms to challenge/attack deep neural network models, a variety of defence mechanisms have been proposed, to improve network robustness or mitigate the issue of facing adversarial perturbations.

Model robustness against adversarial attacks can be addressed during training~\cite{goodfellow2014explaining}, most simply by augmenting the training set with adversarial examples. Goodfellow {\et}~\cite{goodfellow2014explaining} regularized the training of a deep neural network model by an adversarial objective function based on the fast gradient sign method:
\begin{align}
&\hat{J}(\theta,x,y) =\\ & \alpha \cdot J(\theta,x,y) + (1-\alpha) \cdot J(\theta, x + \epsilon \cdot sign(\nabla_x J(\theta,x,y))) \nonumber
\end{align}
\noindent where $J(\cdot)$ is the training objective function. Given the universal approximator theorem\footnote{Universal approximator theorem: A neural network with at least one hidden layer can represent any function to an arbitrary degree of accuracy, assuming the hidden layer to have enough units.}, they argued that if adversarial examples are encoded in the training process, the model should learn those examples and become more robust in dealing with those (and hopefully other) examples.  That is, the model is continually supplied with adversarial examples, such that they essentially resist the current version of the model, applied in an iterative process to make the model progressively more robust. 
While this technique does improve the robustness of deep neural networks, it has been shown by Moosavi-Dezfooli {\et}~\cite{moosavi2017universal} that there is still an effective and yet a universal adversarial example to fool even such adversarially trained networks; nevertheless this approach is still the most common in increasing deep neural network robustness. 

The robustness of deep neural network models can also be improved by simply performing a pre-processing step on the input data. The adversarial attacks are usually generated as an additive perturbation, almost noise~\cite{goodfellow2014explaining}, on the input data. As such, the attack can be defended by heuristically removing input noise, using any number of signal processing techniques, such as a moving average operator, or taking advantage of compression methods~\cite{guo2017countering} to remove high-frequency values. 

Motivated by these ideas, Xie {\et}~\cite{xie2017adversarial} illustrated that random resizing or random padding of input images reduces the effectiveness of adversarial attacks.  At the same time it is worth noting that these changes to the input data may also reduce the accuracy of the model.

Goldblum {\it et al.}~\cite{goldblum2019adversarially} proposed a new distillation-based approach, which incorporates robust training of a student network. The proposed method follows a similar technique as adversarial training, but in the context of distillation, where a ``student'' network is sought to mimic the teacher's outputs within an $\epsilon$-ball of training samples.

It is also possible to take advantage of post-processing techniques to improve the network robustness. Utilizing an ensemble of deep neural networks~\cite{kariyappa2019improving} allows decision-making to be aggregated across several models, improving robustness against adversarial attacks.  This strategy can be combined with averaging~\cite{xie2019feature} or noise perturbation~\cite{Jeddi2019L2P} to make adversarial attacks more difficult.  Table~\ref{tab:defense_summary} summarizes the main techniques to improve robustness of deep neural networks against adversarial attacks.

\begin{table*}[]
    \centering
    \caption{Different defense mechanisms to improve the robustness of deep neural network models against adversarial attacks. These approaches can be performed in pre-processing, during training, post-processing, or even by changing the network architecture.}
    \begin{tabular}{|l|l|c|}
    \hline
        \bf Method  & \bf Procedure & \bf Description      \\ \hline
        Random Resizing/Random Padding~\cite{xie2017adversarial} & Pre-Processing & \makecell{Changing the size of the input image \\ before passing to the network. }    \\ \hline
        Adversarial Training~\cite{goodfellow2014explaining}  &  Training   &   \makecell{Adding targeted perturbed samples\\ to the training data.}   \\ \hline
        Compression~\cite{guo2017countering}  & Pre-Processing    &  \makecell{Compress and decompress the input samples  \\ before passing to the network.}    \\\hline
        Distillation~\cite{goldblum2019adversarially}  & Training/Post-Processing    &   \makecell{Training a student network given \\the original (teacher) network. }      \\ \hline
        Ensemble~\cite{kariyappa2019improving} &  Post-Processing &  \makecell{Aggregating the decision of several networks \\ to mitigate the effect of adversary. }   \\ \hline
        Noise Perturbation~\cite{Jeddi2019L2P} &  Architectural Change/Training   & \makecell{Adding auxiliary noise modules in the network \\ to neuter the effect of adversarial perturbation.}    \\ \hline
    \end{tabular}
    \label{tab:defense_summary}
\end{table*}

\subsubsection{Adversarial Attacks for Autonomous Driving Systems}

Autonomous driving systems are among the pioneering fields in advancing machine learning and deep learning techniques, and complex deep learning networks are utilized in all phases from perception to decision-making and control.  As a result, the effect of adversarial attacks has been significantly investigated for autonomous driving applications. 

Chen {\it et al.}~\cite{chen2018shapeshifter} proposed a new approach to fool the well-known Faster RCNN object detection. They generated adversarial attacks that trick the model into not detecting stop signs in a scene. They took advantage of {\em Expectation over Transformation} \cite{athalye2018obfuscated}, where a random noise is added in each iteration of the optimization to infer more robust adversarial perturbations. Lu {\it et al.}~\cite{lu2017adversarial} proposed an optimization method searching through different sets of stop sign examples to craft new stop sign images which are not detectable by Faster RCNN or YOLO network architectures.

However it has been argued~\cite{lu2017no} that designing adversarial perturbation for real systems, in practice, is harder than those analyzed in the literature and in laboratory environments.  \mbox{Lu {\it et al.}~\cite{lu2017no}} explained that since object detection and decision-making in autonomous driving cars are performed based on a sequence of frames, it is far more difficult to fool a model for all frames, than for one frame. 

\begin{figure}
    \centering
    \includegraphics[width=0.5\textwidth]{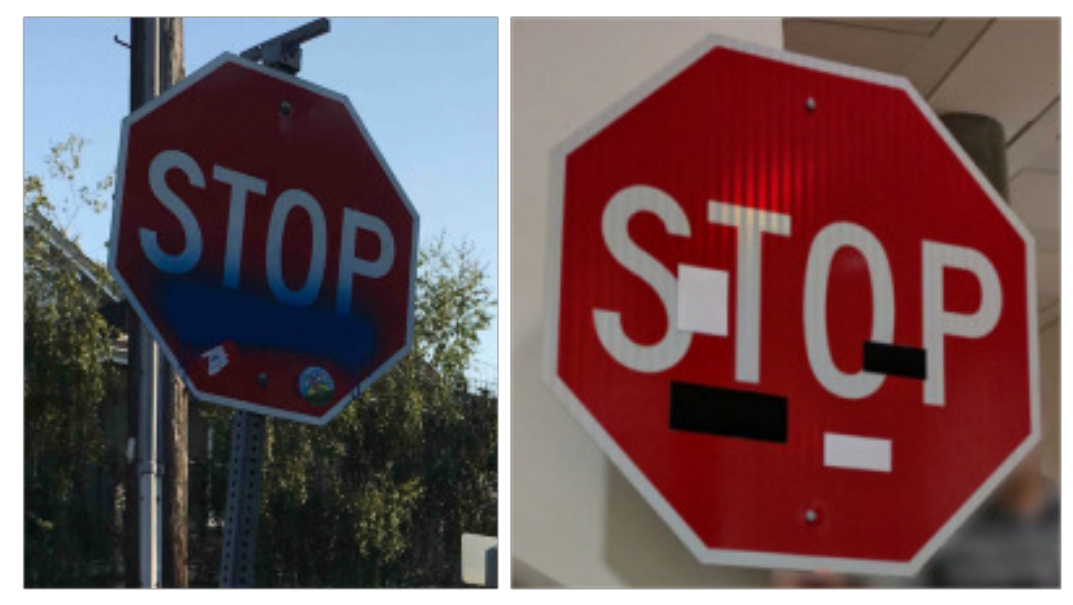}
    \caption{An example of a physical perturbation designed to fool a deep neural network model to misclassify a stop sign. Adding these types of perturbations may not often draw a human's attention as they may be assumed to be regular artifacts on the traffic sign (an example shown in the left image). The right image demonstrates a physical attack that would cause the model to misclassify the stop sign. The example is extracted directly from~\cite{eykholt2017robust}.  }
    \label{fig:physical_attack}
\end{figure}

To be sure, recent algorithms have been proposed to craft adversarial attack in real environments.
Eykholt {\it et al.}~\cite{eykholt2017robust} introduced robust physical perturbation methods which can generate perturbations under different physical conditions.  Figure~\ref{fig:physical_attack} demonstrates an example of this approach, where a physical perturbation is added to a stop sign, causing the model to misclassify the stop sign.    
They took several conditions into account including environmental constraints, spatial constraints and physical limits on imperceptibility while optimizing the perturbation noise. These conditions are incorporated into the loss function during the optimization step. It is worth mentioning that these types of changes to traffic signs are usually considered as ordinary changes which usually do not bring up any suspicions for human inspectors. In other words, these types of perturbations might not be noticeable to human inspection as viable intrusions.

\begin{figure}
    \begin{tabular}{cc}
    \centering
    \includegraphics[width=0.23\textwidth]{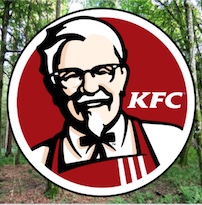} & \includegraphics[width=0.23\textwidth]{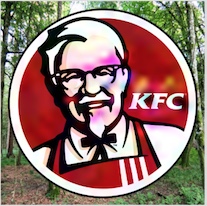} 
\end{tabular}
 \caption{Example of the adversarial attack proposed by~\cite{sitawarin2018rogue}, where logos and advertisements are modified to cause a wrong interpretation by the deep neural network model.  The left image shows an advertisement logo, and the right image shows the modified logo designed to cause a perception model to recognize it as a ``bicycle crossing'' sign. The example is extracted from~\cite{sitawarin2018rogue}.  }
    \label{fig:ad_attack}
\end{figure}

Sitawarin {\it et al.}~\cite{sitawarin2018rogue} proposed a new algorithm which modifies logos and advertisements such that the deep learning model detects them as different traffic signs in a scene. Figure~\ref{fig:ad_attack} shows an example of this scenario. The refinement on the advertisement logo can fool the targeted model to classify the image as a ``Bicycle Crossing'' sign.  

While most of the research in the literature has focused on deep neural networks taking RGB images as input, these issues are not RGB-specific. Recently, Cao {\it et al.}~\cite{cao2019adversarial} analyzed the vulnerability of LiDAR-based methods in autonomous driving systems. They proposed an optimization based approach to generate real-world adversarial objects, evading the LiDAR-based detection framework, generating 3D objects which can fool the network and be invisible to detection.

\section{Best Practices Toward Robust Autonomous Driving}

In the previous section, we discussed the factors and issues influencing the robustness of an autonomous driving system.  In this section, we will discuss the best practices to achieve more reliable systems for autonomous driving applications.

The key part of an autonomous driving system which might challenge the robustness of the system are machine learning models working to navigate the car. As a result, these models should be evaluated thoroughly before any deployment. While the robustness evaluation of each model individually is a first step, the robustness of these models must also be evaluated in a combined, closed-loop framework before deployment, as evaluated by  Tuncali {\it et al.}~\cite{tuncali2018simulation}. The proposed framework is a simulated environment which tries to identify problematic test scenarios by a falsification method. The proposed simulated environment is comprised of four different parts: i) perception system (the machine learning model to be examined), ii) controller, iii) vehicle and environment, and iv) renderer.  The controller takes the detection information from the machine learning model and object class information, and then estimates the actual positions of the objects (e.g., pedestrians or vehicles). The role of vehicle and environment modeling and the renderer is to generate driving scenarios, improving testing and making it closer to real-world situations without deployment.   

\begin{figure*}[!]
    \centering
    \includegraphics[width=1\textwidth]{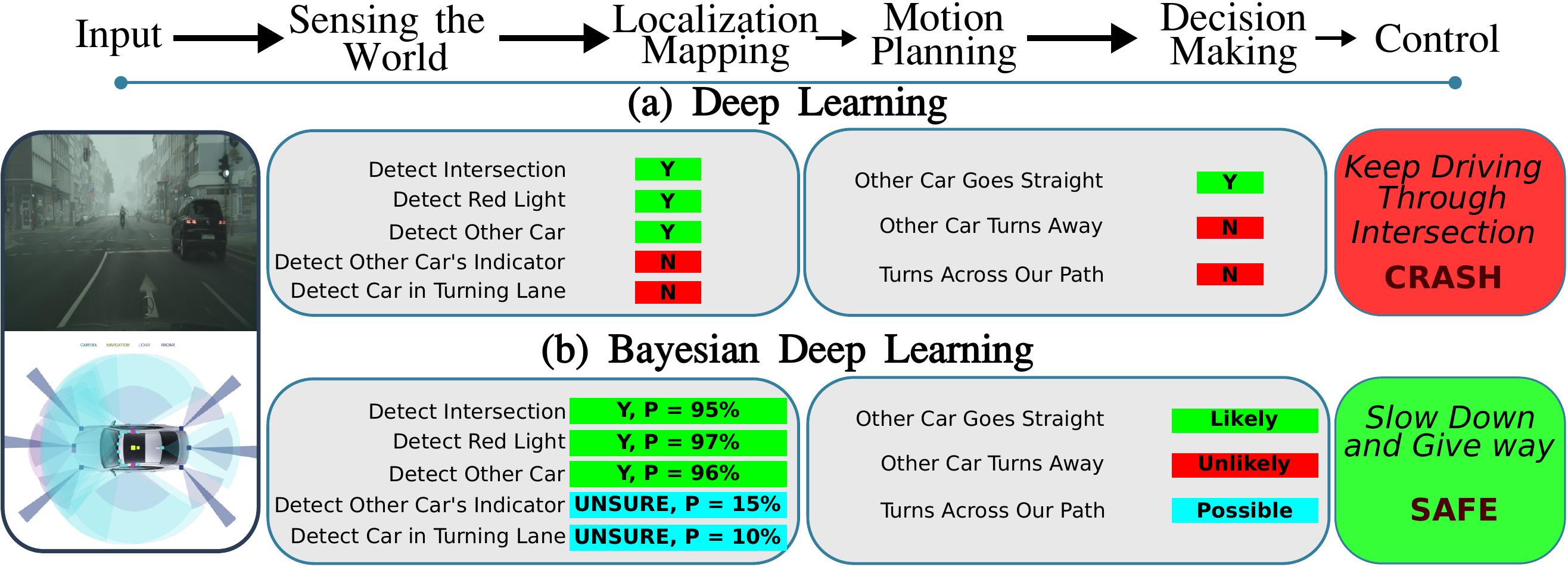}
    \caption{The effect of Bayesian Deep Learning: Model (a) is an end-to-end DNN model with no uncertainty evaluation, while model (b) is Bayesian Deep Learning based,  calculating and propagating uncertainties. In this example, the BDL-based model~\cite{mcallister2017concrete} makes a better decision at the end and prevents a fatal accident. The diagram is drawn following the structure introduced in~\cite{mcallister2017concrete}. }
    \label{fig:uncertainty}
\end{figure*}

Generating test cases leveraging real-world changes in driving conditions like rain, fog, snow and lighting conditions are important in the evaluation of autonomous driving systems. Having the capability of generating real-world conditions would help to identify cases which lead to problematic behaviour by the autonomous system. This approach decreases the need for manual testing of rare-case scenarios to some extent. To this end, Tian {\it et al.} proposed DeepTest ~\cite{tian2018deeptest}, an automated approach to generate samples subject to a wide variety of environmental conditions, to detect erroneous behaviors of deep neural networks models. The proposed framework uses linear and convolutional transformations to change the brightness, contrast or adding fogginess or rain into real images. The idea is that autonomous driving systems should behave similarly for specific scenes with these types of variations; for example, that the steering angle (determined by a learned network) should not change significantly for a given scene, but with different  lighting or weather conditions applied. This testing approach can help to pinpoint corner-cases of inconsistent system behaviour.

There are several research studies  examining the effectiveness of simulated data in improving the accuracy and robustness of perception models in autonomous driving. Ros \mbox{\it et al.}~\cite{ros2016synthia} introduced SYNTHIA, a synthetic dataset of urban scenes to improve the performance of semantic segmentation models in autonomous applications. Their experimental results demonstrated that using synthetic data in conjunction with real-world data can boost the average per-class accuracy. This improvement is significant in classes having limited dataset instances, such as pedestrians, cars and cyclists. Furthermore, it has been stated by different autonomous driving groups, such as Waymo, that taking advantage of simulated driving systems improves the accuracy and performance of models in real-world scenarios and in public roads.

Simulated environments are important in finding corner-cases effectively, and resolving them by providing better training data or more effective learning. However, it is also important to improve the intrinsic robustness of machine learning algorithms. One of the main issues with deep learning models is that they are fairly deterministic, and furthermore a lack of measuring the uncertainty in the decision-making process, making it easy to estimate and predict a network's decision-making and making networks vulnerable to attack.

There have been improvements in deep neural network models to provide uncertainty while making predictions~\cite{ gal2016dropout}. Model uncertainty measures a DNN's confidence associated with a prediction; for example, an autonomous vehicle DNN system could be exposed to test data that has a different distribution from that of the training data (for example trained on urban data and tested in highway environments), and would ideally be expected to be uncertain in its predictions \cite{michelmore2018evaluating}, where uncertainties may be aleatoric (data dependant) or epistemic (model dependant)~\cite{mcallister2017concrete}. Unfamiliar test data distributions, as a result of either insufficient original training data or a distributional shift, are common examples causing model uncertainty~\cite{shafaei2018uncertainty}.

Gal~\cite{gal2016uncertainty} argued that Softmax  outputs in a DNN classifier should not be interpreted as its prediction confidence. They experimentally showed that for out-of-distribution test samples, a model can have high Softmax outputs at times when predictions should be highly uncertain. Moreover, Gal and Ghahramani~\cite{gal2016dropout} illustrated that by using stochastic regularization techniques, DNNs can be viewed as Bayesian approximators of Gaussian process. Using this Bayesian Deep Learning (BDL) view, one can easily estimate a model's confidence without needing to change its architecture.

For end-to-end autonomous systems in which a DNN takes raw sensory data as input and maps them to controlling commands (e.g., steering, braking, acceleration, etc.), finding model uncertainties is a simple procedure. \mbox{Michelmore {\it et al.}~\cite{michelmore2018evaluating}} proposed to add dropout layers to NVIDIA's PilotNet (i.e., an end-to-end self-driving car system) as a stochastic regularizer in the training step. However, they were also utilized at test time to extract the model confidence, by computing the model output multiple times for each input image.

Figure~\ref{fig:uncertainty} shows an illustration of two different end-to-end autonomous systems with and without model uncertainty. Bayesian deep neural networks can be substituted to provide uncertainty in the decision-making process. As seen in Figure~\ref{fig:uncertainty}, by incorporating uncertainty in the decision-making process and taking advantage of probabilistic approaches, the system can provide more reliable predictions, better decision-making, and safer actions. 

However, in a modularized autonomous architecture, where the system is organized as a pipeline of subsystems, evaluating model uncertainties is a much bigger challenge than that of end-to-end architectures. McAllister {\it et al.}~\cite{mcallister2017concrete} stated that in order to prevent errors generated in perception subsystems to not propagate through the rest of the pipeline and affect the entire decision making process, all of the subsystems should be equipped with Bayesian Deep Learning tools, to allow uncertainty distributions to be taken into and propagated by each subsystem.

\section{Conclusion}

Autonomous driving offers potentially major advantages to society, such as reducing injury, decreased insurance costs, and reductions in gasoline usage.  The past few years have witnessed remarkably significant progress towards fully automated vehicles being present on public roads. However there do still remain concerns regarding the reliability of the computer vision and data analysis models operating within autonomous vehicles, and even more significantly their robustness in different situations. In this manuscript, we examined the robustness of autonomous systems, focusing on the   proper functioning in adverse conditions/environments and in the presence of intrusions and adversarial attacks. Practical solutions to mitigate these issues and to improve the robustness of these models were discussed, ranging from extending datasets by simulated data, simulated evaluation environments to uncover corner cases, and new techniques to better calculate the uncertainty of such models in decision-making, all strategies which can help to improve the performance of models in real-world applications.

The tremendous success of autonomous driving has opened a vast range of opportunities, for researchers across a wide range of domains, but also for members of society beginning to imagine a different future.  This excitement has led to raised expectations and optimistic timelines about how soon such vehicles might be expected, however for reasons of safety and engineering ethics, it is essential to fully understand the robustness and reliability of the designed systems.


\input{Main_Final_Reduced_Ref.bbl}

\pagebreak

{\huge Supplementary Material} \\ 

This part provides further details and an extended literature review on the topics discussed in the main manuscript.

\setcounter{section}{0}
\renewcommand\thesection{\Alph{section}}

\section{Autonomous Driving Systems}
The general architecture of an autonomous driving system~\cite{paden2016survey,badue2019self} consists of
\begin{enumerate}
    \item {\bf Sensing the world,} consisting of various sensors and algorithms processing the available data and providing a semantic scene understanding~\cite{cheng2011autonomous,janai2017computer}.
    
    \item {\bf Localization and mapping,} computing the AV pose (location and orientation) with respect to the surrounding environment, which is frequently addressed via simultaneous localization and mapping (SLAM) techniques~\cite{dissanayake2001solution,engel2015large}, well-known in robotics and off-line pre-mapping~\cite{levinson2007map,levinson2010robust}.
    
    \item {\bf Motion planning,} providing different trajectories as sequences of states, given the environment information, initial states, and the final goal~\cite{frazzoli2002real,gonzalez2015review}.
    
    \item {\bf Decision-making,} selecting the optimal trajectory, while considering other factors like safety, compliance, etc.~\cite{paden2016survey,maurer1996compact}.
    
    \item {\bf Control,} actuating components and ensuring that the AV follows the selected path~\cite{paden2016survey,gruyer2015persee}.
\end{enumerate}
\section{Object Detection \& Semantic Segmentation }
Different perception models may be used independently for different tasks such as lane segmentation, traffic sign identification, and traffic light detection~\cite{yurtsever2019survey}. However, they can be formulated as an object detection or a semantic segmentation problem. 

Tables~\ref{table:Cityscape},~\ref{table:Kitti} evaluate the performance of state-of-the-art methods for semantic segmentation, and Table~\ref{table:Kitti-Detection-Car} the performance object detection, based on two well-known autonomous driving datasets:  CITYSCAPE~\cite{cordts2016cityscapes} and KITTI~\cite{geiger2013vision}. For Table~\ref{table:Kitti-Detection-Car}, the object detection accuracy is evaluated only based on detecting cars in the scene; the division into Easy, Moderate and Hard are predetermined categories in the KITTI dataset.

\begin{table}[b]
\centering
\caption{
Semantic segmentation evaluation:  state-of-the-art methods are evaluated based on the predicted mIoU on the CITYSCAPES~\cite{cordts2016cityscapes} test set.}
\begin{tabular}{|l|c|}
\hline
\bf Method & \bf IoU (\%)\\
\hline
ResNet38~\cite{wu2019wider}&80.60\\
\hline
DeepLabV3+~\cite{chen2018encoder}&82.10\\
\hline
DRN-CRL~\cite{zhuang2018dense}  &82.80\\
\hline
Zhu et al.~\cite{zhu2019improving}&83.50\\
\hline
Panoptic-DeepLab~\cite{cheng2019panoptic}&84.20\\
\hline

\end{tabular}
\label{table:Cityscape}
\end{table}

\begin{table}[t]
\centering
\caption{
Semantic segmentation evaluation:  state-of-the-art methods are evaluated based on the predicted mIoU on the KITTI~\cite{geiger2013vision} test set.}
\begin{tabular}{|l|c|}
\hline
\bf Method &\bf IoU (\%)\\
\hline
APMoE Seg~\cite{kong2018pixel}&47.96\\
\hline
SegStereo~\cite{yang2018segstereo}  &59.10\\
\hline
AHiSS~\cite{meletis2018training}&61.24\\
\hline
LDN2~\cite{kreso2017ladder}&63.51\\
\hline
MapillaryAI~\cite{rota2018place}&69.56\\
\hline
Zhu et al.~\cite{zhu2019improving}&72.83\\
\hline

\end{tabular}
\label{table:Kitti}
\end{table}

\begin{table}[t]
\centering
\caption{
3D Detection results over the KITTI~\cite{geiger2013vision} dataset; the state-of-the-art methods are compared based on mean Average Precision (mAP) in detecting 3D boxes for cars only.}
\begin{tabular}{|l|c|c|c|}
\hline
\bf Method &\bf Easy(\%) &\bf Moderate(\%) &\bf Hard (\%)\\
\hline
AVOD~\cite{ku2018joint}&84.41&74.44 &68.65\\
\hline
VoxelNet~\cite{zhou2018voxelnet}&81.97& 65.46&62.85\\
\hline
SECOND~\cite{yan2018second}&87.43&76.48& 69.10\\
\hline
F-PointNet~\cite{qi2018frustum}&83.76&70.92&63.65\\
\hline
PointRCNN~\cite{shi2019pointrcnn}&88.88& 78.63&77.38\\
\hline
STD~\cite{yang2019std}&89.70&79.80&79.30\\
\hline

\end{tabular}
\label{table:Kitti-Detection-Car}
\end{table}

\section{Adversarial Attacks \& Defence Techniques}
Adversarial attacks can be categorized as black-box or white-box attacks \cite{kurakin2016adversarial}. Several techniques have recently been proposed to analyze this issue and to illustrate the drawbacks of deep neural networks. These methods can be grouped together based on the way they calculate the perturbation to be added to the input data. Table~\ref{tab:adversarial_attack_summary} shows some of the most common adversarial attacks and their characteristics.    

Several methods have been proposed to address the issue of deep neural networks facing adversarial attacks. These methods can be applied in different stages of networks training, from pre-processing to post-processing or even changing the architecture of the model. Table~\ref{tab:defense_summary} summarizes some of the common approaches to improve the robustness of deep neural networks. 

\begin{table*}
    \centering
    \caption{Summary characteristics of common adversarial attacks.  Attacks are characterized based on whether they are white-box or black-box (i.e., having accessing to model details or not), the number of network queries to be able to generate the perturbed input (attack frequency), and how they measure the amount of perturbation added to the input data. These characteristics are described based on the most common approaches for which these methods are used, however it is possible to extend these methods to change the characteristics as well. }
    \begin{tabular}{|l|c|c|c|c|}
    \hline
        \bf Method &\bf White-Box/Black-Box & \bf Targeted/Non-Targeted & \bf Attack Frequency  &\bf Measurement  \\ \hline
        FGSM~\cite{szegedy2013intriguing} &  White-Box  & Non-Targeted  &     One-time  & element-wise \\ \hline
        DeepFool~\cite{moosavi2016deepfool} &  White-Box & Non-Targeted   &    Iterative &   $l_p$  \\ \hline
        C\&W~\cite{carlini2017towards}  &  White-Box    & Targeted    &Iterative &  $l_1,l_2,l_\infty$   \\ \hline
        PGD~\cite{madry2017towards}  & White-Box       &   Both  &Iterative &   $l_\infty$  \\\hline
        ZOO~\cite{chen2017zoo}  & Black-Box & Both   &Iterative &    $l_2$ \\ \hline
        One-Pixel~\cite{su2019one} & Black-Box      & Both    &Iterative &   $l_0$  \\ \hline
    \end{tabular}
    \label{tab:adversarial_attack_summary}
\end{table*}

\begin{table*}[]
    \centering
    \caption{Different defense mechanisms to improve the robustness of deep neural network models against adversarial attacks. These approaches can be performed in pre-processing, during training, post-processing, or even by changing the network architecture.}
    \begin{tabular}{|l|l|c|}
    \hline
        \bf Method  & \bf Procedure & \bf Description      \\ \hline
        Random Resizing/Random Padding~\cite{xie2017adversarial} & Pre-Processing & \makecell{Changing the size of the input image \\ before passing to the network. }    \\ \hline
        Adversarial Training~\cite{goodfellow2014explaining}  &  Training   &   \makecell{Adding targeted perturbed samples\\ to the training data.}   \\ \hline
        Compression~\cite{guo2017countering,dziugaite2016study,das2017keeping}  & Pre-Processing    &  \makecell{Compressing and decompressing the input samples  \\ before passing to the network.}    \\\hline
        Distillation~\cite{goldblum2019adversarially}  & Training/Post-Processing    &   \makecell{Training a student network given \\the original (teacher) network. }      \\ \hline
        Ensemble~\cite{kariyappa2019improving,pang2019improving} &  Post-Processing &  \makecell{Aggregating the decisions of several networks \\ to mitigate the effect of an adversary. }   \\ \hline
        Noise Perturbation~\cite{he2019parametric} &  Architectural Change/Training   & \makecell{Adding auxiliary noise modules in the network \\ to neutralize adversarial perturbation.}    \\ \hline
    \end{tabular}
    \label{tab:defense_summary}
\end{table*}

\section{More Detail on Model Uncertainty}
Estimating the uncertainty of deep neural networks is challenging, however several techniques have been proposed to provide uncertainty in deep neural networks while making predictions~\cite{mcallister2017concrete, gal2016dropout, michelmore2018evaluating, huang2019assessing}. Model uncertainty is important factor in the decision-making as it can measure a network's prediction confidence, a confidence which can help to provide a more seamless interaction between human and machine, and to increase the human trust in understanding and interpreting network decisions.

\input{supp.bbl}


\end{document}